# An AI based Digital Score of Tumour-Immune Microenvironment Predicts Benefit to Maintenance Immunotherapy in Advanced Oesophagogastric Adenocarcinoma

Quoc Dang Vu[1], Caroline Fong[2], Anderley Gordon[2], Tom Lund[2], Tatiany L Silveira [2], Daniel Rodrigues[2], Katharina von Loga[2], Shan E Ahmed Raza[1], David Cunningham[2,*], Nasir Rajpoot[1, 3, 4, *]

[1]Tissue Image Analytics Centre, University of Warwick, Coventry, United Kingdom
[2]The Royal Marsden Hospital NHS Foundation Trust and The Institute of Cancer Reseach
[3]The Alan Turing Institute, London, United Kingdom
[4]Histofy Ltd, Birmingham, United Kingdom
*Joint senior authors

# Abstract

Gastric and oesophageal (OG) cancers are the leading causes of cancer mortality worldwide. In OG cancers, recent studies have showed that PDL1 immune checkpoint inhibitors (ICI) in combination with chemotherapy improves patient survival. However, our understanding of the tumour immune microenvironment in OG cancers remains limited. In this study, we interrogate multiplex immunofluorescence (mIF) images taken from patients with advanced Oesophagogastric Adenocarcinoma (OGA) who received first-line fluoropyrimidine and platinum-based chemotherapy in the PLATFORM trial (NCT02678182) to predict the efficacy of the treatment and to explore the biological basis of patients responding to maintenance durvalumab (PDL1 inhibitor). Our proposed Artificial Intelligence (AI) based marker successfully identified responder from non-responder ($p < 0.05$) as well as those who could potentially benefit from ICI with statistical significance ($p < 0.05$) for both progression free and overall survival. Our findings suggest that T cells that express FOXP3 seem to heavily influence the patient treatment response and survival outcome. We also observed that higher levels of $CD8^+PD1^+$ cells are consistently linked to poor prognosis for both OS and PFS, regardless of ICI.

# Introduction

Gastric and oesophageal cancers are a global health burden, representing the fifth and seventh most common cancers globally but fourth and sixth leading cause of cancer mortality [1]. Adenocarcinomas are the most common histological type of oesophagogastric (OG) cancers followed by squamous cell carcinomas. Adenocarcinomas are further subdivided according to anatomical location into types I, 1 to 5cm from the cardia, type II, arising at the cardia, and type 3, 1 to 2cm below the cardia. This anatomical classification scheme has staging and management implications. Outcomes are highly dependent on staging and high stage disease is uniformly and swiftly fatal. TNM staging [2]. [3]Recently, the advent of immune-checkpoint blockade therapy has led to improved outcomes in patients with these malignancies and investigation of immune related biomarkers has become a key field of investigation [4-6]. In this regard, leveraging artificial intelligence (AI) could potentially complement existing clinical and pathological parameters and aid as predictive biomarkers for immune checkpoint inhibitor (ICI) therapy.

ICI has emerged as a new powerful therapeutic treatment in many cancer settings [4-6]. Inhibitors targeting programmed death ligand-1 (PDL1) or programmed cell death protein-1 (PD1) have been shown to be effective in treating renal cell carcinoma, melanoma and non-small cell lung cancer [6]Dissecting the role of different immune cells in the range of human tumours amenable to ICI therapy is a complex task. In colorectal cancer (CRC), immune scores have emerged as strong prognostic markers that are independent from TNM staging [7]-[8]. Specifically, abundance of $CD8^+$ lymphocytic infiltration has been shown to be associated with good prognosis [9]. Quantification of $CD3^+$ and $CD8^+$ lymphocytes at the edge (invasive margin) and at the core of the tumour could also allow planning for follow-up strategies [8-10].

Several studies have attempted to further investigate the role of immune cell subtypes in gastric cancer (GC). However, immune activity within the GC tumour immune microenvironment (TiME) is more complex than in CRC. For instance, while the abundance of memory T cells in CRC is linked to better prognosis in colorectal and stomach cancers, it leads to worse survival in oesophageal cancer (EC) [11]. Recent findings about $FOXP3^+$ are contradictory [12-15]. In particular, although cells expressing FOXP3 are typically immunosuppressive [15] and thus their abundance is commonly associated with poor survival [12,13]. *Huang et al.* [14] has shown that a high amount of $FOXP3^+$ infiltration could lead to better outcome in GC. It is worth noting that while FOXP3 is primarily expressed by immunosuppressive regulatory cells ($CD4^+FOXP3^+$), it can also be expressed on other cell types, such as Natural Killer (NK) [16] or Gamma Delta (γδ) T cells [17,18]. As a result, the presence of $FOXP3^+$ cells in GC and EC may have important prognostic significance [19-21].

[4-6]In oesophagogastric adenocarcinoma (OGA), data from recent practice changing trials showed that PDL1 inhibitors in combination with chemotherapy was associated with overall response rates of 58%−62% in all patients [22,23]. As ICI directly targets the immune system dynamics within the TiME, computational approaches for robust quantification of T-cell subtypes within these patients could potentially lead to further understanding of the immune system and better prediction of response to the maintenance immunotherapy.

In parallel, artificial intelligence (AI) and machine learning (ML) methods have recently been developed to assist with objective and reproducible quantification of cancer phenotypes and/or predicting treatment outcomes in various cases [24]. AI/ML methods have also been proposed to find association between the density of various cell types and molecular pathways [25] and prediction of Gleason Grade in prostate cancer [26]. If designed and implemented carefully, AI/ML methods can be used to derive interpretable features that are prognostically relevant and carry potential to help in precise quantification of interactions of immune cells in the TiME.

Here, we analysed multiplex immunofluorescence (mIF) images of full-face tumour sections from patients with advanced OGA who received first-line fluoropyrimidine and platinum-based chemotherapy in the PLATFORM trial (NCT02678182) with an AI based analytical pipeline to interrogate the biological basis of response to maintenance durvalumab (PDL1 inhibitor) and predict its efficacy. We propose an AI based digital score of the OGA TiME that can better stratify patients receiving maintenance durvalumab into responders versus non-responders. In addition, we seek to determine whether using the same digital score we can identify patients in the surveillance arm who could have potentially benefited from receiving the same ICI treatment that was given to patients in the treatment arm. Our findings suggest that the proposed score has the potential to improve our understanding of mechanisms behind immunotherapy and develop better treatment strategies for patients with advanced OGA.

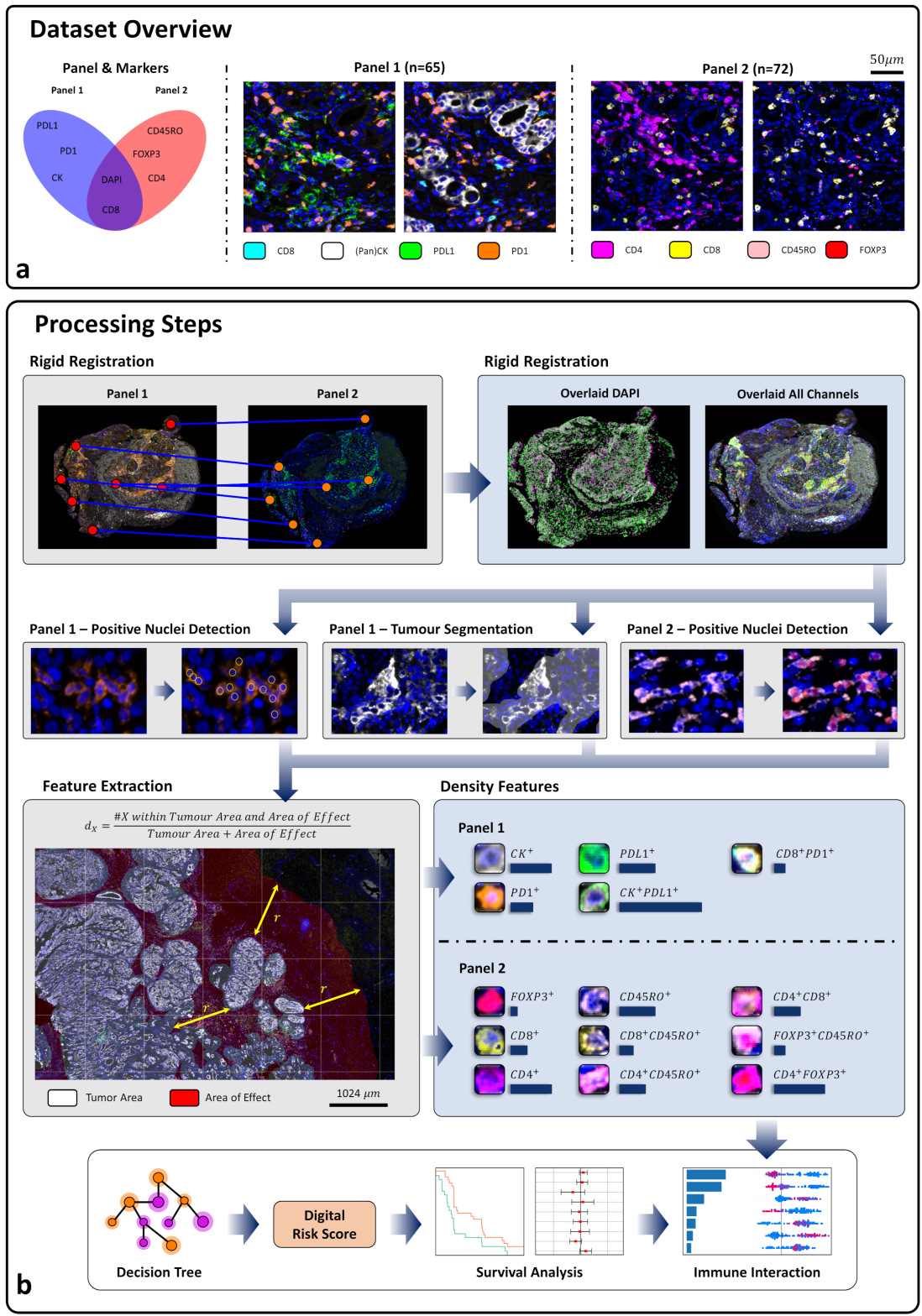

**Fig. 1 Data overview and processing pipeline. a.** The dataset was collected using two mIF panels, where each panel used different set of markers as mentioned above. **b.** The WSIs from both panels were aligned, and positive nuclei detection for each marker was performed based on the registered WSIs. The phenotype of each nucleus was then extracted by combining its positivity across markers. Density features were later calculated for each nuclear phenotype based on the tumour area and its neighbourhood. Based on these density features, a Digital Risk Score (DRS) was generated for each patient using a tree-based method, with higher scores indicating worse survival. The scores were later utilised to stratify patients with respect to their survival time, and the effect of each feature on the Digital Risk Score was investigated using a post-hoc method.

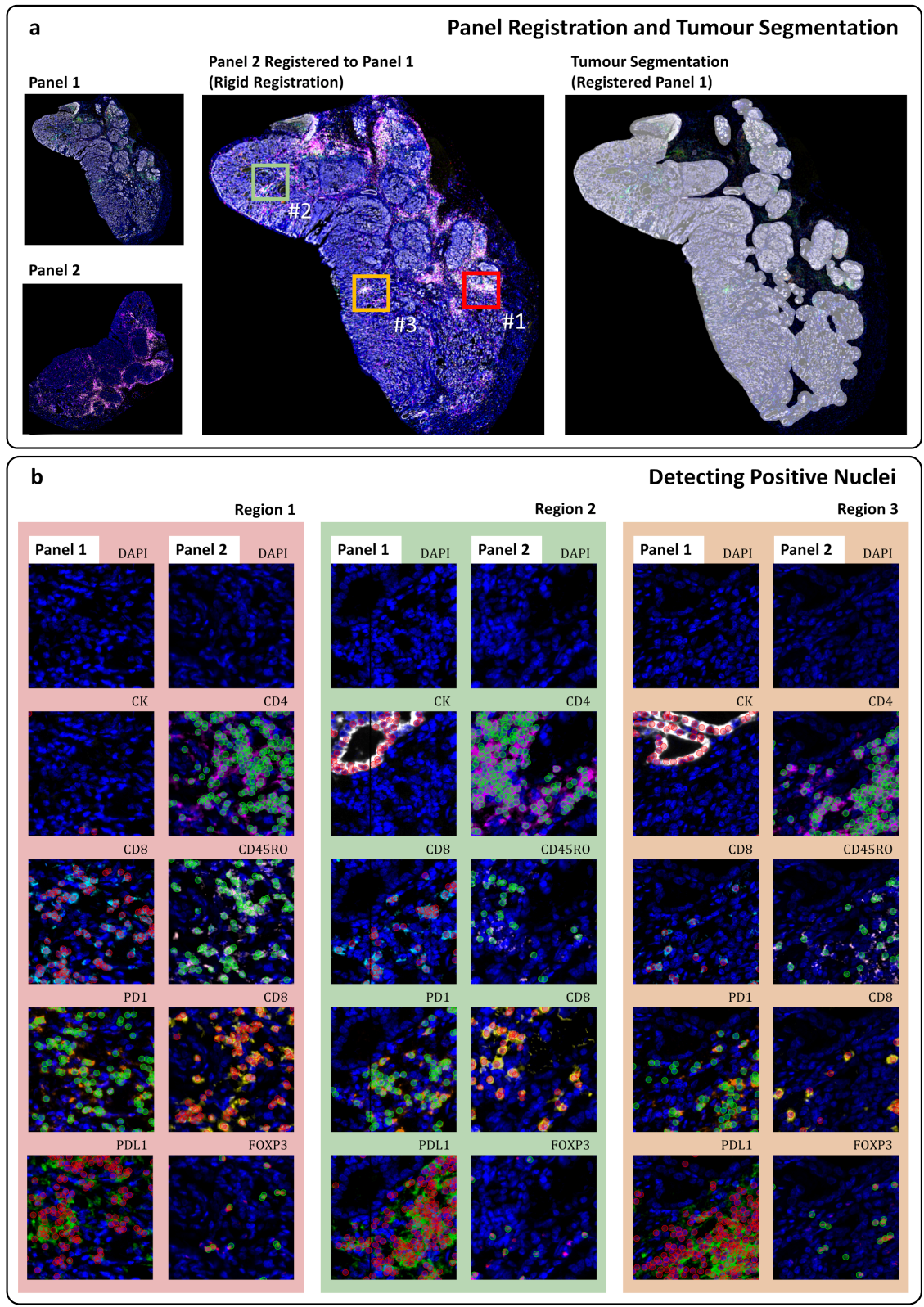

**Fig. 2 Sample results for registration, segmentation and nuclei detection. a.** Two WSIs from different sections of a same patient, one was stained for Panel 1 and the other was for Panel 2. WSI for Panel 2 was registered to the one for Panel 1 using a rigid transformation. Then, tumour segmentation was conducted on the WSI from Panel 1; **b.** 3 areas within the registered WSI were selected to demonstrate the results of detecting positive nuclei for each marker. The detected nuclei were marked as green and red circles for visualization purposes. The DAPI channel was left empty for easier tracing same nuclei across multiple markers. As for the staining colour (such as white or orange), please refer to **Fig. 3**a for details.

**Fig. 4 Phenotypes utilised in the study, their expressions and detected sample cell nuclei.** For each nucleus, the positivity detected for each marker from **Fig. 1** and **Fig. 2** were combined to derive its actual phenotype. Because nuclei in Panel 1 and Panel 2 are not the same due to them being from non-adjacent tissue sections, phenotypes obtained from Panel 1 may express markers that exist in Panel 2 or vice versa.

# Results

## Dataset overview

Clinical data were collected from patients with advanced OGA who achieved radiological response or disease control following first-line platinum-based chemotherapy and were randomised to active surveillance or maintenance systemic therapies, including the PDL1 antibody durvalumab. Our dataset consisted of 65 patients in the active surveillance arm (ARM1) and 72 patients in the durvalumab arm (ARM3). Following disease progression, patients were followed up until death or withdrawal of consent. From this dataset, we observed that patients in ARM1 have a median survival time of approximately 4 months for progression free survival (PFS) and 10 months for overall survival (OS), whereas those in ARM3 have a median survival time of around 3 months for PFS and 10 months for OS. Additional clinical details about the patients' pathological characteristics are provided in the **Supplementary**.

**Multiplex immunofluorescent (mIF) staining**

Prior to the treatment, multiple tissue samples were collected from patients and digitally scanned using the Vectra Polaris system at a resolution of $0.25 \mu m$/pixel to generate mIF whole-slide images (WSIs). Before scanning, tissue sections were stained with two panels: Panel 1 included Pan-Cytokeratin (CK), DAPI, CD8, PD1, and PDL1, while Panel 2 included DAPI, CD4, CD45RO, CD8, and FOXP3. For this study, we ensured that each patient had at least one WSI from each panel. For this purpose, 3um sections were cut and adjacent sections were stained.

**Workflow and feature construction**

The markers utilised in each panel for staining the mIF WSIs as well as the workflow of this study are summarised in **Fig. 1**. For each patient, we employed an automated key-point matching method to register Panel 2 WSIs to their corresponding Panel 1 WSIs using the Panel 1 WSI as a reference image. Additional details on the registration process are included in the **Methods Section**. After confirming the registered WSIs were of sufficient quality through qualitative and quantitative analyses, we used a variant of U-Net [27] on the image channels containing DAPI (nuclei counter stain) and CK (a tumour positive membrane marker) to perform tumour segmentation on Panel 1 WSIs. **Fig. 2a** shows sample results of the WSI registration and subsequent tumour segmentation, with white-coloured segmented tumour areas overlaid on top of the registered WSI from Panel 1.

As a next step, we analysed the positivity of nuclei for different types of markers (PD1, PDL1, CD4, CD8, CD45RO, CK and FOXP3) in the registered WSIs. We used an in-house state-of-the-art deep learning method [28] to locate nuclei in the registered WSIs, and then used a ResNet50 model to determine whether nuclei were positive for a given marker. Since the registered WSIs are composed of WSIs from non-adjacent tissue sections, we thus have two sets of detected nuclei for each registered WSI, one for each panel. Furthermore, because each panel has four markers, for a given location within the registered WSI, there are eight sets of positivity predictions. We provide additional details regarding this process in the **Methods Section**. We show results of detected positive nuclei for three sample regions in **Fig. 2b**.

Registration results at a higher resolution level are also shown in **Fig. 2b**. It can be observed that the overall gland morphology remains almost intact as we could see the two glandular segments are located less than $5 \mu m$ apart. This indicates qualitatively that the registration procedure achieved satisfactory alignment; for instance, correspondence between the two panels can be observed by visual examination of the spatial distribution of nuclei expressing CD8 in Region 1. A quantitative evaluation of the performance of our cross-WSI registration method has been provided in the **Methods Section**. We note that the correspondence offset is within the confines of the image patch (around $128 \mu m$). This offset could be further minimised by using adjacent tissue sections.

Using the segmented tumour area and the positive nuclei detected per marker as described above, we built an analytical pipeline for extracting WSI-level density features for each cell phenotype. We determined the phenotype of each cell based on its positivity across all markers within the same panel. We combined $CK^+$ cells within the tumour area to obtain tumour cells. In **Fig. 3**, we list 14 cell phenotypes analysed in this study along with their detected samples and their abbreviations for subsequent discussion. The density of each cell phenotype was calculated for the tumour area and its vicinity within a 128μm extent.

Finally, using 14 quantitative density features (based on the formular described in **Fig. 1**), one for each cell phenotype mentioned in **Fig. 3**, we utilised a tree-based ML model by the name of XGBoost to calculate a Digital Risk Score (DRS) for each patient for subsequent response prediction and survival based risk stratification. This approach was chosen to account for possible non-linear interactions between cell phenotypes. The final set of DRSs was obtained using the standard train/validation/testing (or the train/validation/intra-arm validation subsets) settings in machine learning, and the predictions were aggregated from multiple random splits to provide a less biased DRS estimation as detailed in the **Methods Section**. In addition, we also investigated how changes in the density of each phenotype impact the resulting DRS.

**DRS identifies responders to immunotherapy**

We assessed the ability of DRS values to identify patients who are likely to respond to ICI treatment. For this experiment, we exclusively utilised data from ARM3 (treatment arm) to compute the DRSs. Patient DRSs from the testing subsets of multiple random splits (referred to as *intra-arm validation*) were collated, resulting in a single DRS for each patient in ARM3. The DRS values were used to divide the patients in the unseen testing subsets into two groups: the High Risk group (non-responders) and the Low Risk group (responders). We computed the threshold for separating these two groups based on the DRSs from the validation subsets of all corresponding random splits. A detailed description of this process can be found in the **Methods Section**.

We plotted the Kaplan-Meier (KM) curves in **Fig. 4a** to evaluate the survival rates of the two groups and compared the significance of group separation using the log-rank test. From the figure, we observe that patients allocated to the Low Risk group according to the proposed DRS had better survival rates (~20% and ~40%) compared to those in the High Risk group (~5%) for both PFS (censored at 12 months, $p = 0.001$) and OS (censored at 36 months, $p = 0.003$). Interestingly, the survival probability of High Risk group decreased after a certain period, while the Low Risk group remained relatively stable. Specifically, the survival probabilities of the High Risk and Low Risk groups were approximately similar for the first 3 months of PFS (~65%) and the first 6 months of OS (~90%). However, by the 6th month in PFS and the 18th month in OS, the survival probabilities of patients in High Risk group declined to ~20% in both categories. In contrast, the survival probabilities of patients in Low Risk group only declined to 50% and 40% respectively. By the 12$^{th}$ month for PFS and the 36$^{th}$ month for OS, patients in the Low Risk group had survival probabilities of around 20% and 35%, respectively, while those in the High Risk group had only about a 5% probability of survival.

**DRS identifies patients who could potentially benefit from immunotherapy**

To identify patients who could potentially benefit from immunotherapy, we employed a similar approach as in the previous section. We first trained a model on ARM1 (the surveillance arm) patients only and generated DRSs for patients from the *intra-arm validation* of ARM1. We then applied the same model to unseen patients in ARM3 (the treatment arm) and obtained their DRSs, termed as *cross-arm validation*. Similar to the previous section, using a single threshold derived from DRSs of the validation splits within ARM1, we stratified ARM1 intra-arm validation DRSs and ARM3 cross-arm validation DRSs into Low Risk and High Risk groups. Intuitively, patients in ARM1 who may benefit from the ARM3 treatment would show up as either ARM3 High Risk (A3HR) with better survival than ARM1 High Risk (A1HR) or as ARM3 Low Risk (A3LR) with better survival than ARM1 Low Risk (A1LR).

From the KM curves for A1HR and A1LR groups shown in **Fig. 4b**, we observe that patients in A1LR had better survival than those in A1HR. This was statistically significant for both PFS censored at 12 months ($p = 0.001$) and OS censored at 36 months ($p = 0.0001$). This is a positive affirmation that the DRSs generated from the model trained only with ARM1 could effectively stratify patients in the unseen testing subsets of ARM1 based on the DRS.

We then plot KM curves to compare the survival of A1HR and A3HR, as well as of A1LR and A3LR in **Fig. 4c**. In terms of PFS, there was no significant difference in survival for A1LR and A3LR group ($p > 0.05$). However, patients in A3HR had a better survival than those in A1HR ($p = 0.01467$). For the first ~3 months, A1HR had a similar survival rate (reaching ~60%) as A3HR. However, by the 6th month, the survival rate in A3HR decreased to ~30% whereas the survival rate in A1HR reached ~10%. By the 12-month mark, the survival rate in A3HR group declined to ~20% while A1HR reached ~2%.

Our analysis of the OS also found that A1LR and A3LR groups had a similar survival rate ($p > 0.05$). However, patients in A3HR group had a significantly better survival than those in A1HR group ($p = 0.02348$). Both these groups initially had a similar survival rate, reaching around 40% by the 12-month mark, before diverging significantly until the end of the 36-month period. By the end of the study period, the survival rate of patients in A3HR declined to around 20% whereas the survival rate of those in A1HR was barely above 0%.

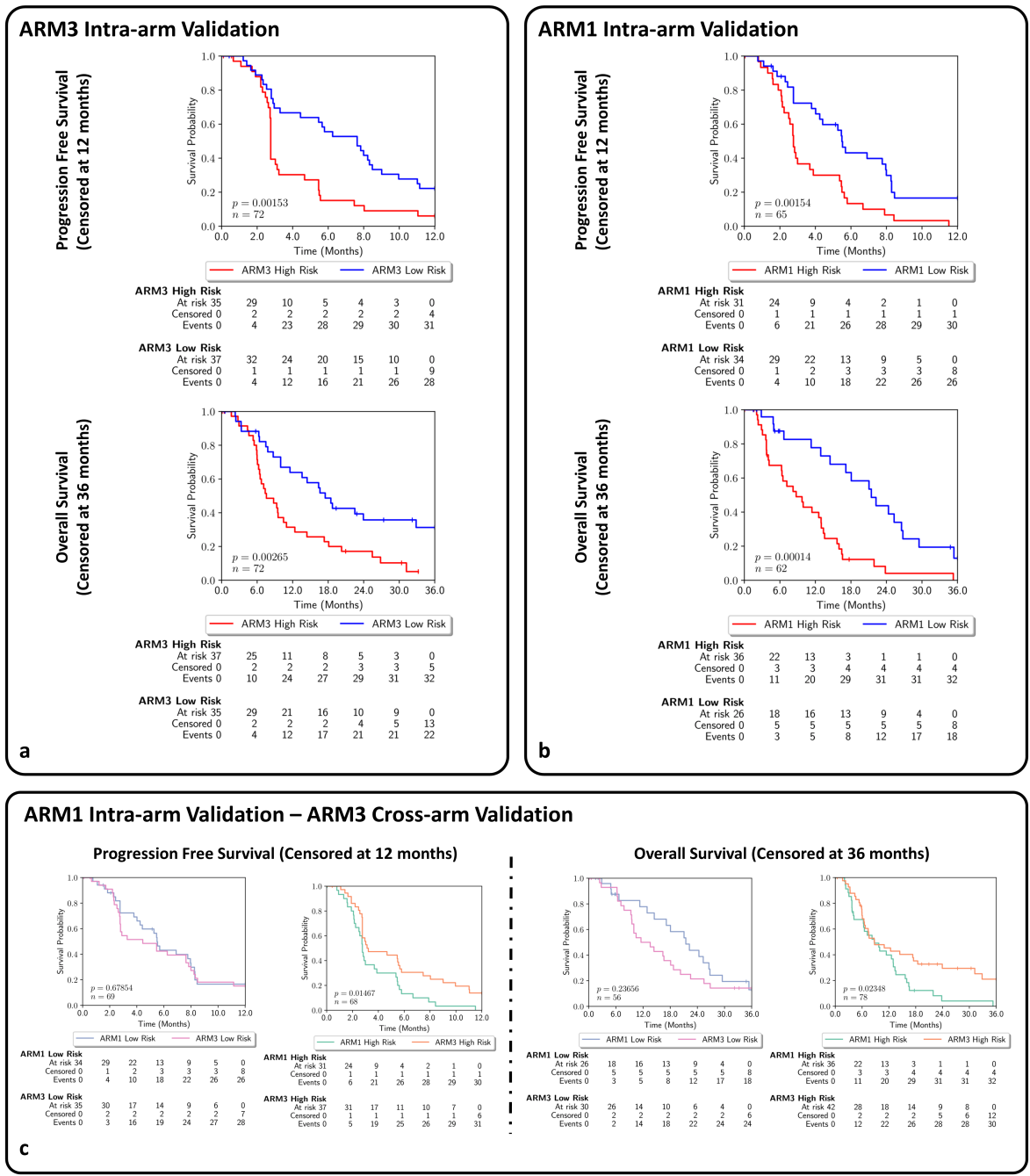

**Fig. 5 Kaplan-Meier (KM) survival curves for ARM1 and ARM3.** Digital Risk Scores (DRS) were calculated by combining the prediction on the test portions of multiple random data splits, with higher scores indicating worse survival. Patients were stratified into High Risk or Low Risk groups using these scores, with the threshold for such separation is the average of the median DRS obtained from the validation portion of all random splits; **a.** DRS were obtained using a tree-based method trained exclusively on data from ARM3. They successfully separated patients in ARM3 into responders (Low Risk) and non-responders (High Risk) to immunotherapy with significant differences in both PFS and OS (log-rank test $p \ll 0.005$). **b.** DRS were obtained using the same tree-based method but trained only on data from ARM1, which also effectively stratified patients into Low Risk and High Risk groups with significant differences in both PFS and OS (log-rank test $p \ll 0.005$); **c.** The same procedure as in (b) was repeated on unseen data from ARM3 (cross-arm validation data), resulting in a new set of High Risk and Low Risk groups. By comparing the KM curves for patients in ARM1 and ARM3, we found that patients in the High Risk group from ARM1 could potentially benefit from immunotherapy, as evidenced by better survival in the ARM3 High Risk group compared to the ARM1 High Risk group. Notably, this benefit became apparent only after approximately 3 months for PFS and 12 months for OS.

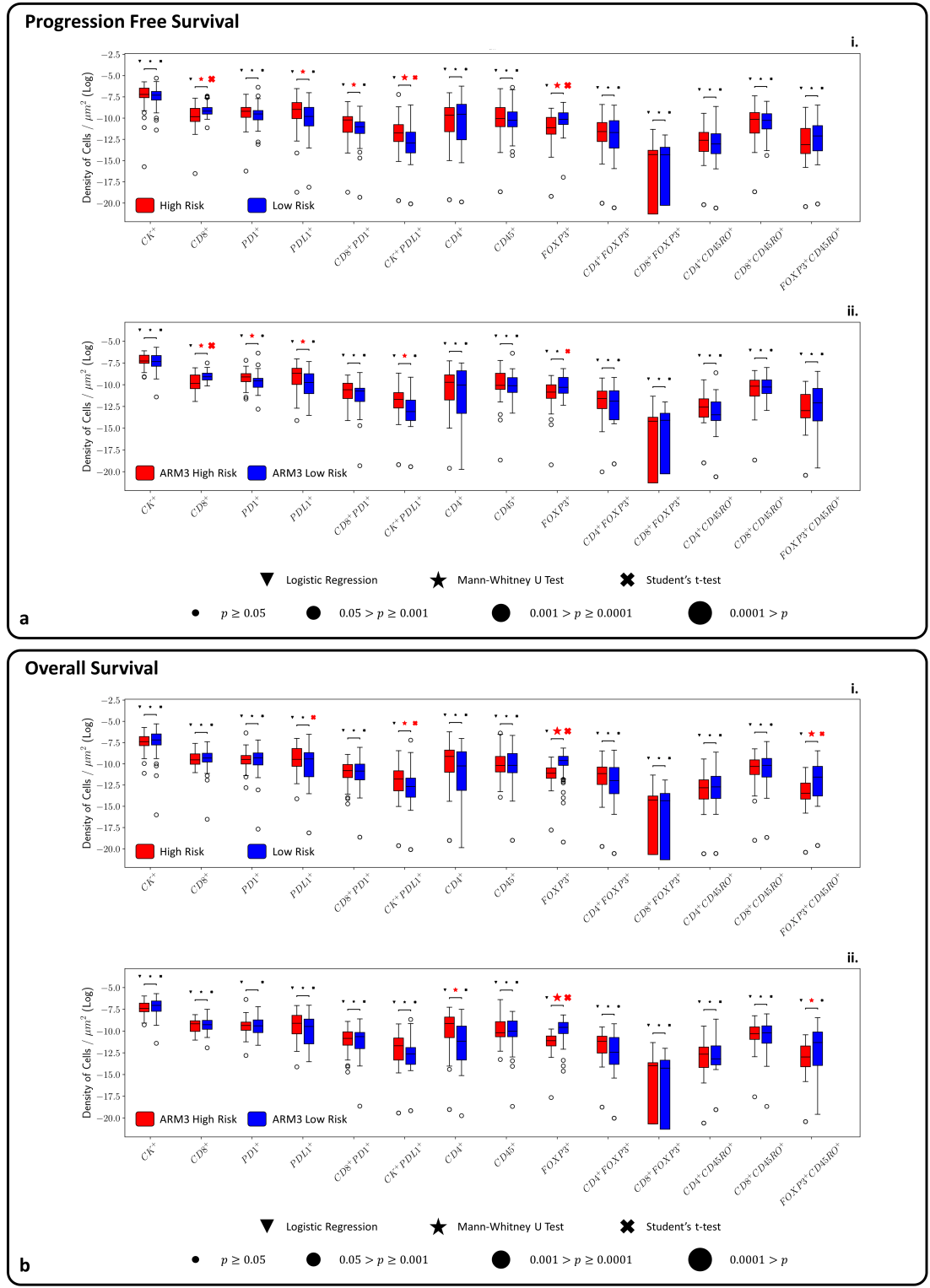

**Fig. 6 Difference in immune subtypes between identified patient groups.** Statistical analysis was conducted to compare the density of each cell phenotype between the different groups identified in **Fig. 4a**. Mann-Whitney U test and Student's t-test were utilised. We also conducted Logistic Regression test to explore the possibility of linearly separating the two groups under investigation using only the cell density of that phenotype. If the statistical difference between the groups is $p < 0.05$, the test symbols are highlighted in red; **a.** Comparison of Low Risk versus High Risk within ARM3 alone (i) and over the entire dataset (ii) for PFS; **b.** Comparison of Low Risk versus High Risk within ARM3 alone (i) and over the entire dataset (ii) for OS.

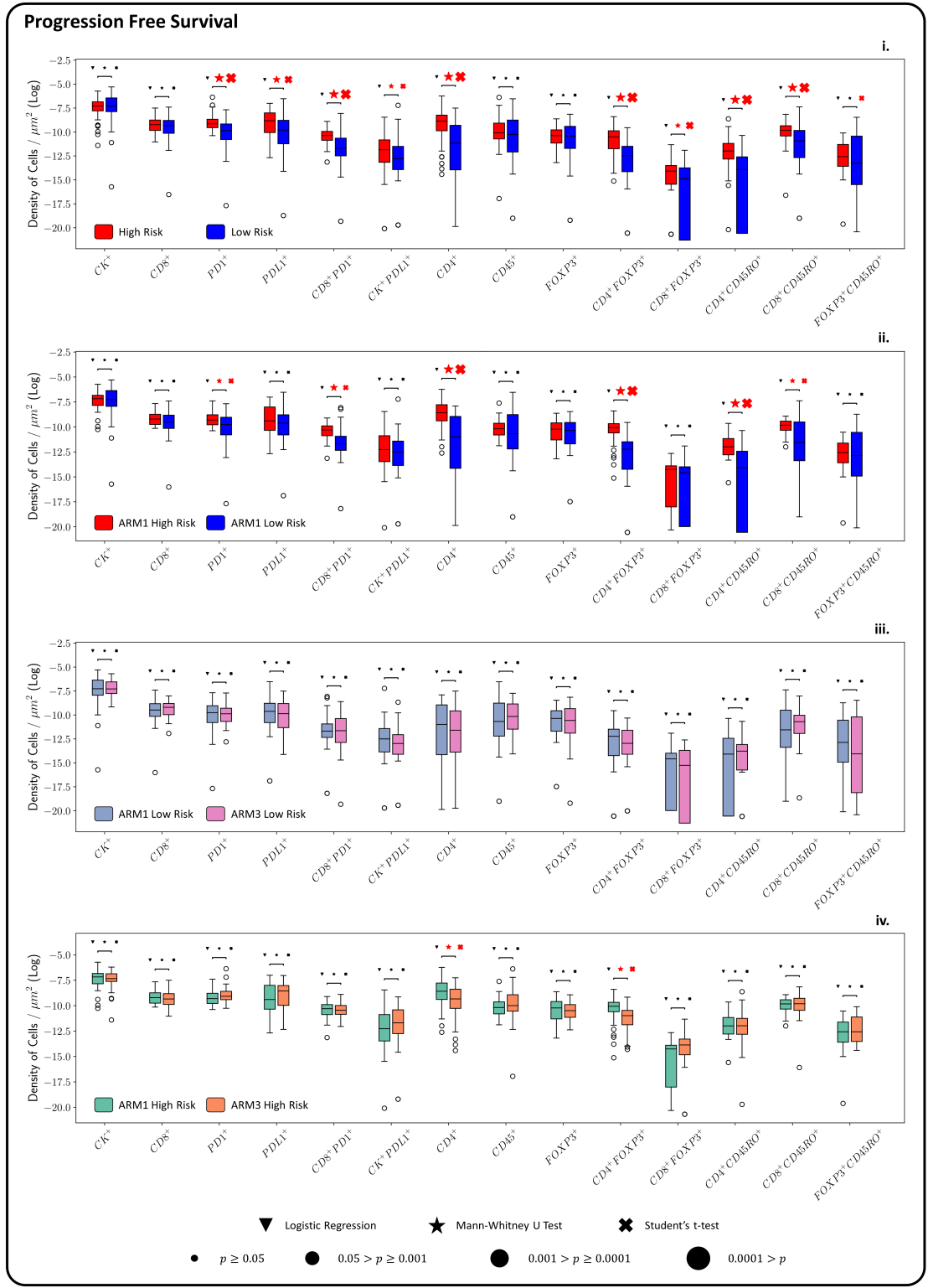

**Fig. 7 Difference in the immune system between identified patient groups.** Statistical analysis was conducted to compare the density of each cell phenotype between the different groups identified in **Fig. 4b** and **Fig. 4c**. Mann-Whitney U test and Student's t-test were utilised. We also conducted Logistic Regression test to explore the possibility of linearly separating the two groups under investigation using only the cell density of that phenotype. If the statistical difference between the groups is $p < 0.05$, the test symbols are highlighted in red. We compared the Low Risk versus High Risk within ARM1 alone (ii) or over the entire dataset (i) for PFS. Additionally, the cell density within ARM1 Low Risk was compared with ARM3 Low Risk (iii), as well as those within ARM1 High Risk against ARM3 High Risk (iv).

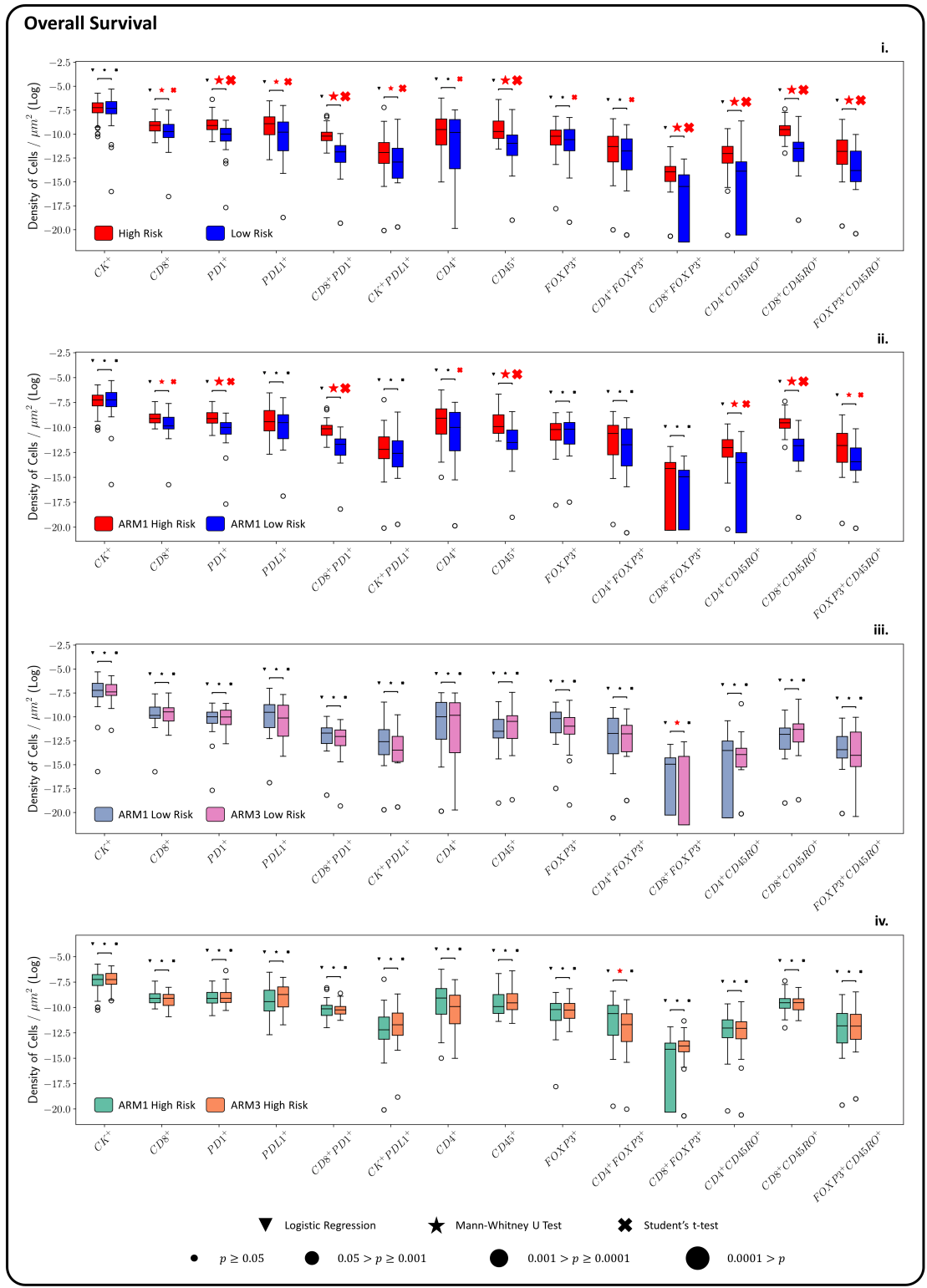

**Fig. 8 Difference in the immune system between identified patient groups.** Statistical analysis was conducted to compare the density of each cell phenotype between the different groups identified in **Fig. 4b** and **Fig. 4c**. Mann-Whitney U test and Student's t-test were utilised. We also conducted Logistic Regression test to explore the possibility of linearly separating the two groups under investigation using only the cell density of that phenotype. If the statistical difference between the groups is $p < 0.05$, the test symbols are highlighted in red. We compared the Low Risk versus High Risk within ARM1 alone (ii) or over the entire dataset (i) for OS. Additionally, the cell density within ARM1 Low Risk was compared with ARM3 Low Risk (iii), as well as those within ARM1 High Risk against ARM3 High Risk (iv).

**Differences in the TiME between identified groups of patients**

We investigated potential differences in the TiME between patient groups identified from previous sections. We used Mann-Whitney U and Student-t tests to compare the distribution of the cell density features, one for each of the 14 phenotypes (as shown in **Fig. 3**), and tested if each feature could directly distinguish between patient groups using logistic tests.

From **Fig. 5a-ii** and **Fig. 5b-ii**, within ARM3, we observe that responders to ICI in terms of both OS and PFS had a significantly higher density of FOXP3$^+$ cells as compared to non-responders ($p < 0.001$ for Student-t test). For PFS, we additionally see that responders had a higher density of CD8$^+$ but a lower density of PD1$^+$, PDL1$^+$ and CK$^+$PDL1$^+$ as compared to non-responders ($p < 0.05$ for Mann-Whiteney U test). As for OS, responders have lower density of CD4$^+$ but higher FOXP3$^+$CD45RO$^+$ ($p < 0.05$ for Mann-Whiteney U test). We further extended such stratification to also include ARM1 and plotted the comparisons in **Fig. 5a-i** and **Fig. 5b-i**. When observed over the entire population, the difference in FOXP3$^+$ became even more pronounced ($p < 0.0001$ for Mann-Whiteney U test) for both OS and PFS. For PFS, we observed similar differences between the two groups for CD8$^+$, PDL1$^+$ and CK$^+$PDL1$^+$ ($p < 0.05$ for Mann-Whiteney U test). However, the density of CD8$^+$PD1$^+$ in non-responders was statistically higher ($p < 0.05$ for Mann-Whiteney U test) than that in responders. As for OS, the difference in FOXP3$^+$CD45RO$^+$ remains unchanged ($p < 0.05$ for Mann-Whiteney U test) whereas CD4$^+$ is no longer statistically different. However, we additionally observed that non-responders have significantly higher PDL1$^+$ and CK$^+$ PDL1$^+$ than responders ($p < 0.05$ for Student-t test).

In **Fig. 7** and **Fig. 8**, we present the findings of our TiME analysis for patient groups identified in the previous section. From **Fig. 7-iii** and **Fig. 8-iii**, aside from CD8$^+$FOXP3$^+$, there is virtually no difference in the TiME between patients in A1LR and A3LR for both OS and PFS ($p > 0.05$). Meanwhile, for PFS, **Fig. 7-iv** shows that CD4$^+$ and CD4$^+$FOXP3$^+$ in A1HR are higher than A3HR ($p < 0.05$ for Mann-Whitney U test). As for OS, from **Fig. 8-iv**, A1HR only differs from A3HR in CD4$^+$FOXP3$^+$ in which it has higher density ($p < 0.05$ for Mann-Whitney U test).

On the entire population, for PFS, **Fig. 7-i** shows that patients in the High Risk group have significantly higher density ($p < 0.05$) of CD4$^+$, CD4$^+$FOXP3$^+$, CD4$^+$CD45RO$^+$, CD8$^+$CD45RO$^+$, CD8$^+$PD1$^+$, PD1$^+$, PDL1$^+$ and CK$^+$PDL1$^+$ than those in the Low Risk groups. In contrast, for OS, aside from CK$^+$, **Fig. 8-i** shows that patients in the High Risk group have higher density in all remaining phenotypes ($p < 0.05$).

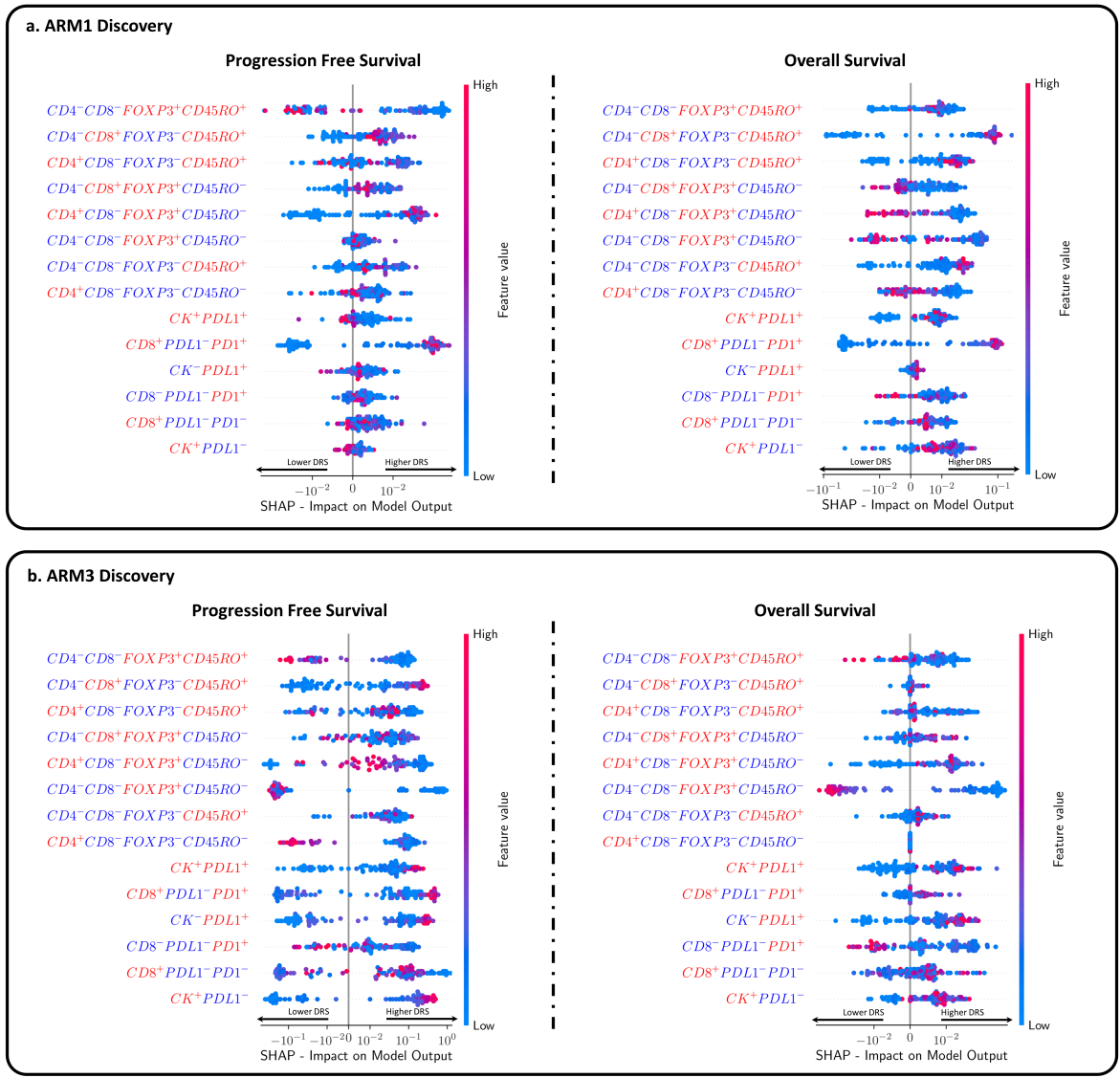

**Fig. 9 How each feature affects the resulting Digital Risk Score (DRS) on sample-level and population-level.** We present the SHAP plot for the models and DRSs utilised in **Fig. 4**. Each dot in each feature row represents a data point, which corresponds to a patient. The colour of each dot indicates the magnitude of that patient feature compared to the entire population, with red indicating higher and blue indicating lower values. The position of each dot on the x-axis reflects the contribution of the feature to the corresponding patient's DRS, with dots to the left indicating a lower DRS (better survival) and dots to the right indicating a higher DRS (worse survival). However, dots that are closer to the vertical line in the middle of the plot have less of an impact on generating the DRS. **a.** SHAP plot for the models utilised in **Fig. 4b**, which were trained only on ARM1 data (termed ARM1 Discovery), and its corresponding DRSs for the entire dataset; **b.** SHAP plot for the models utilised in **Fig. 4a**, which were trained only on ARM3 data (termed ARM3 Discovery), and its corresponding DRSs for the entire dataset.

### How densities of cell phenotypes influence the Digital Risk Scores

Considering the models trained solely on ARM1 data as ARM1 Discovery (those utilised in **Fig. 4**b) and the models trained only on ARM3 data as ARM3 Discovery (those utilised in **Fig. 4**a), we performed SHAP [29] analysis on the ARM1 Discovery and ARM3 Discovery models, as well as their corresponding DRSs from the entire population (including both the test (intra-arm validation) portions and unseen data of each respective model) in **Fig. 9**. Additional details on how the results in **Fig. 9** being interpreted are provided in **Supplementary S1**.

Based on **Fig. 9**a and **Fig. 6**b, the association between each phenotype density and DRS can be observed. In the surveillance arm (ARM 1), elevated levels of $CD8^+CD45RO^+$ typically lead to higher DRS in both PFS and OS. In particular, for PFS, increased levels of $CD8^+PD1^+$ are also commonly associated with higher DRS. Conversely, in OS, heightened levels of $CD4^+CD45RO^+$, $CD45RO^+$, $CK^+PDL1^+$ or $CK^+$ generally lead to a higher DRS, while elevated levels of $FOXP3^+$ or $CD4^+$ T cells overall are associated with a lower DRS. Interestingly, higher levels of $FOXP3^+CD45RO^+$ produce lower DRS in PFS but higher DRS in OS, whereas this association is reversed for higher levels of $CD8^+FOXP3^+$ or $CD4^+FOXP3^+$.

In patients who received maintenance ICI (ARM 3), higher levels of $CD4^+FOXP3^+$, $CK^+PDL1^+$, $CD8^+PD1^+$, $PDL1^+$ or $CK^+$ cells are generally associated with a higher DRS in both PFS and OS. Conversely, higher levels of $FOXP3^+$ or $PD1^+$ typically lower DRS in both PFS and OS. In PFS specifically, elevated levels of $CD8^+CD45RO^+$, $CD4^+CD45RO^+$, $PDL1^+$ or $CD8^+$ are generally linked to higher DRS, while higher levels of $FOXP3^+ CD45RO^+$ or $CD4^+$ tend to decrease DRS value. The influence of $CD4^+FOXP3^+$ is particularly mixed as a lower amount of $CD4^+FOXP3^+$ could strongly lead to either a higher DRS or a lower DRS. This manifests as two unicolor blue clusters on the left most and right most side of the plot.

## Discussion

In contrast to traditional methods, like PCR or CIBERSORT[30], which rely on gene analysis or other bioinformatic approaches to infer quantitative and qualitative information about the TiME, our approach utilises multiplexed immunofluorescence images, providing a cost-effective alternative. This image-based analysis not only allows for the quantification of the general distribution of T cell populations but also facilitates the exploration of their intricate spatial relationships across entire tissue samples. This stands in stark contrast to conventional gene analysis techniques, which often limit insights to specific regions of interest and are unable to disaggregate the contributions of distinct cell types, thus potentially failing to capture the full complexity of a patient's TiME [31].

Moreover, our approach incorporates cutting-edge machine learning techniques for the rapid and robust quantification as well as identification of individual T cells and their phenotypes, presenting a significant acceleration compared to traditional assessment methods. Additionally, the subsequent derivation of DRS based on these identified T cells further enhances our ability to model and comprehend the dynamics of T cell populations within the TiME. The capacity of our approach for extracting deep insights about the TiME goes beyond conventional analyses that solely assess patient outcomes, such as OS and PFS, based on isolated T cell populations.

The proposed DRS reveals that higher levels of $CD4^+FOXP3^+$ are generally associated with poor survival for OS and/or PFS, regardless of ICI. These findings are consistent with the existing literature [32]. We also observed that higher levels of $CD8^+PD1^+$ are consistently linked to poor prognosis for both OS and PFS, regardless of ICI. [33,34] , CD8+PD1+ is associated with immunosuppression and the link between PD1+ and response to anti-PD-L1 therapies is very poorly characterised. Extended immunosupresion of these cells is thought to trigger a terminal differentive state known as a TEMRA. This cell state is terminally suppressed and therefore non responsive to any ICI interaction. Further work would need to define the TEMRA T cell state within this disease setting.[35]In patients receiving ICI, we found that $FOXP3^+$ as an independent biomarker, strongly associated with better survival in both OS and PFS. This is consistent with work published previously in gastroesophageal cancer [19]. Here, we report a more substantial analysis and have found that the FOXP3+ population is also double negative (CD4$^-$CD8$^-$). Coupled with our limited data, unravelling the precise nature of this enigmatic population stands as a promising avenue for future research.

We also observe that the abundance of CD4$^+$CD45RO$^+$ or CD8$^+$CD45RO$^+$ is associated with poorer survival outcomes independent of ICI. The expression of CD45RO$^+$ in this experimental design identifies a population of T cells that have seen antigen (known as effector T cells) before transitioning into long term stable immunological memory cells. [11,36] The abundance of CD45RO$^+$ T cells is associated with favourable prognosis in oral squamous, lung, prostate and colorectal cancer [37]. The indication that CD45RO$^+$ is associated with poorer survival outcomes in the OGA setting is a novel finding that will require more in depth experimental development and mechanistic information to elucidate this finding [38]. Despite these findings, there are a few limitations. Our work relies heavily on machine learning techniques, which are known to be data hungry. Our dataset is small and may lead to noisy observations and reporting. However, we have provided robust statistical analysis to address this problem as much as possible with accessible data. Our statistics are measured on biopsy samples, which may not entirely reflect the true heterogeneity of tumour tissue [31].

In summary, we proposed a digital risk score (DRS) using non-linear modelling and advanced machine learning techniques for assessing the benefit of maintenance immunotherapy in patients with advanced OGA. Our findings suggest that patients with higher DRS often have elevated levels of nearly all cell phenotypes compared to those with lower DRS, and that the existing TiME can greatly affect patient survival with or without ICI. While our observations raise intriguing questions, further validation is necessary to fully understand their role in the TiME. Further investigations using better staining techniques, full resection samples, larger cohorts and automated methods are required before these findings can be implemented in clinical practice.

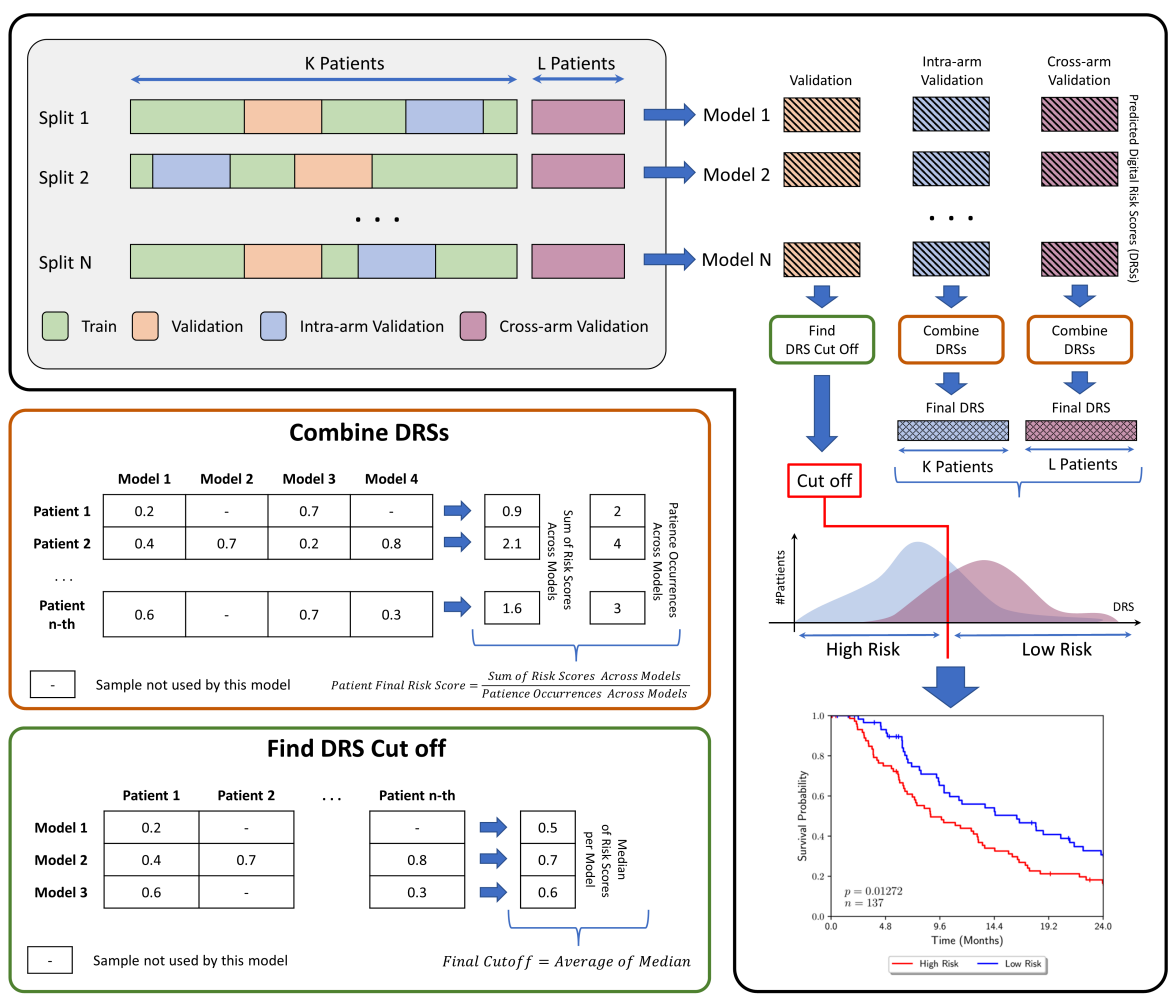

**Extended Data Fig. 1. Summary on how Digital Risk Scores (DRSs) were obtained in the study.** Depending on the experiment, patients within ARM1 or ARM3 were randomly selected 25 times (i.e., 25 data splits) into train/validation/testing (or train/validation/intra-arm validation) portions (60/20/20). Train portions were utilised for training the tree-based model while the validation portions were used to select the final set of models. Predictions of the final models were combined and subsequently utilised for stratifying patient survival.

**Extended Data Table 1. Performance of the final tree-based models utilised in the study.** Reported results are mean ± standard deviation of C-Index measured across 25 random data splits.

|  | ARM1 Discovery | | ARM3 Discovery | |
| --- | --- | --- | --- | --- |
|  | **OS** | **PFS** | **OS** | **PFS** |
| **Validation** | 0.7401±0.0751 | 0.6985±0.0816 | 0.6876±0.0811 | 0.7468±0.0631 |
| **Intra-arm Validation** | 0.7514±0.0661 | 0.7101±0.0745 | 0.6793±0.0841 | 0.7648±0.0658 |

# Methods

**Rigid image registration**

To register whole slide images (WSIs) from different panels, we first manually matched tissue chunks within each WSI to each other across panels, resulting in 647 tissue pieces from 137 pairs of WSIs. These tissue pieces varied in size, ranging from 3000×3000 to 30000×30000 pixels at 0.5µm resolution. We then rigidly registered each pair of these tissue pieces using key points extracted from their DAPI channels at $4\mu m$ or $8\mu m$ resolution.

We employed R2D2 [39], which utilised a CNN, to robustly extract multi-scale key points from each DAPI image. R2D2 key points are optimised for robustness, repeatability, and efficiency, and have been shown to outperform traditional key point detection algorithms like SIFT, SURF and ORB [40] in various image registration tasks, including medical image registration, remote sensing, and microscopy. R2D2 key points are also known for their superior speed, accuracy, and robustness to changes in illumination, noise, or occlusion compared to traditional methods. After extracting the R2D2 key points from the DAPI images, we employed RANSAC (Random Sample Consensus) to reliably obtain matching key points and subsequent image transformation.

Finally, we evaluated this registration procedure performance using Pearson Correlation Coefficient (PCC) and Structural Similarity Index (SSIM) at multiple resolutions. We measured the PCC and SSIM of 647 registered tissue pieces and respectively reported their mean ± standard deviation across the entire dataset. Our procedure respectively achieved SSIM and PPC of 0.5591±0.1034 and 0.3817±0.1283 at 1.0µm; 0.5236±0.1045 and 0.4011±0.1314 at 2.0µm; and 0.4895±0.1077 and 0.4459±0.1380 at 4.0µm. With these results, we consider our registration achieved sufficient quality for subsequent analysis.

**Tumour segmentation**

We developed a deep learning model for image segmentation that classifies each pixel within an image as normal or tumour using only DAPI and CK channels. Our model is based on a U-Net-based CNN [27] with ConvNext [41] as the encoder. To leverage the pretraining of the encoder in RGB colour space, we ordered the channels as {DAPI, CK, DAPI}.

To train the model, we used 35 whole slide images (WSIs) that were pixel-wise annotated by pathologists. We used 20 WSIs for training and reserved 10 WSIs for validation and model selection. From each WSI used for training, we extracted patches of 2048x2048 pixels and with a stride of 512×512 at 0.5µm resolution. Patches that contained less than 25% background were retained for usage.

The performance of the final model was evaluated using the DICE (Sørensen–Dice) [42] in which our model achieved a value of 0.8628. This indicates that our model is capable of accurately classifying normal and tumour pixels in mIF images using only DAPI and CK channels.

**Nuclei positivity detection and classification**

**Nuclei detection**

In our study, we employed a robust approach for nuclei detection within whole slide images (WSIs) using only the DAPI channel. Specifically, we utilised the HoVer-Net [28] architecture with ConvNext [41] as the encoder, and manually annotated 36 image tiles to train and validate our model. These image tiles were extracted from 28 WSIs, and the annotations were verified by an expert pathologist, resulting in a total of 122110 nuclei being annotated. The size of the image tiles ranged from 512×512 to 2048×2048 pixels at 0.5μm resolution.

To train our model, we selected 6 image tiles from 4 WSIs, resulting in 92713 nuclei being used for training. The remaining 30 image tiles, with a total of 29397 nuclei, were reserved for validation and model selection. We evaluated the performance of our instance segmentation model using three metrics: Panoptic Quality (PQ), Detection Quality (DQ), and Segmentation Quality (SQ) [43]. Our final model achieved PQ, DQ, and SQ scores of 0.4902, 0.7260, and 0.6723, respectively.

Furthermore, we also assessed the performance of our model in nuclei detection alone and achieved an F1 score of 0.7871. For this F1 score calculation, we used the Hungarian algorithm to uniquely pair the centroids of detected nuclei with those of the ground truth nuclei. Pairs with a distance larger than 3μm were considered as failed pairings, and the corresponding detected and ground truth nuclei were classified as False Positive and False Negative, respectively.

With these results, we deem our nuclei detection (localization) achieved sufficient performance for subsequent analyses.

**Classifying nuclei positivity**

To classify nuclei as positive or negative with respect to a given marker, we employed a pretrained ConvNext [41]. For each nucleus identified using the method above, we extracted an image patch of size 256×256 at 0.25μm resolution centring on the nucleus centroid for each of their corresponding marker channels. Each of these patches then have their channels ordered as {DAPI, MARKER, MASK}, where MARKER denotes the channel for the current marker and MASK contains the instance pixels identified by the HoVer-Net.

We annotated 6305, 4723, 3400, 3739, 3839, and 6666 image patches as positive for CD4, CD8, CD45RO, FOXP3, PD1, and PDL1, respectively, across 8 WSIs. From these positive image patches, we queried the nearest neighbour to obtain 135216 negative image patches.

For training, we used nuclei patches from 4 WSIs (18177 positive, 91851 negative) and reserved the remaining nuclei patches (10495 positive, 43365 negative) for model validation and selection. We evaluated the performance of the model in terms of Average Precision (AP) for each marker and selected the final model based on the highest mean AP (mAP) across all markers. The selected model achieved an overall mAP of 0.918 and individual AP of 0.957, 0.841, 0.920, 0.974, 0.919, and 0.898 for CD4, CD45RO, CD8, FOXP3, PD1, and PDL1, respectively. It is worth noting that aside from FOXP3, which stains the nuclei content, the other markers stain the cell transmembrane, making it unnecessary to specifically train for CK among the eight markers.

Overall, our 2-stage approach using CNNs for nuclei localization and then marker classification provides a robust and accurate method for analysing histopathological images.

**Tree-based method and Digital Risk Scores**

**Tree-based method training overview**

The features in **Fig. 3** were utilised as input to XGBoost [44] for estimating each patient risk with respect to an event. We denote this digital assessment in the paper as Digital Risk Score (DRS). XGBoost is a specific implementation of gradient boosting for decision tree. Where the predictions from multiple weak learners (decision trees) are additively combined to "boost" the final prediction model. This boosting model can be roughly defined as:

$$f(X) = \Sigma_{m=1}^{M} \beta_m g_m(X_m, \theta_m)$$

Given a set of input samples $X \in \mathbb{R}^{N \times D}$ consisting of $N$ samples, each with $D$ features, their predictions are generated as a weighted sum of $M$ decision trees. Each decision tree, denoted as $g_m$, is trained using a subset $X_m$ of the original data and/or features, which can be represented as $\{X_m \in X; X_m \in R^{L \times K}; L \leq N; K \leq D\}$. Here, $L$ denotes the number of samples and $K$ denotes the number of features selected out from $X$ to make $X_m$. As each decision tree is trained independently, their respective $\theta_m$ and $X_m$ may differ from each other. Finally, the contribution of each decision tree to the final prediction is determined by a weighting term denoted as $\beta_m$.

In this study, we optimised the function described above for assessing a patient survival using the partial likelihood of the Cox Proportional Hazard (CoxPH) model[45]. Thus, the resulting DRSs can be loosely interpreted as Hazard Ratios compared to a baseline hazard function. However, unlike the traditional CoxPH model, the baseline hazard function is typically not retrievable in XGBoost formulation. Consequently, it is also important to note that DRS is not a probability of an event occurring for a patient and DRSs from different models cannot be directly compared. Nevertheless, within a single model, higher DRS values indicate a higher likelihood of an event occurring, and vice versa (higher DRS values imply worse prognosis).

**Evaluation and utilization of Digital Risk Score**

Our study adopted a strategy similar to that employed in a previous study[46]. We defined the discovery cohort as having three parts: train, validation, and intra-arm validation (or testing set) portions, all of which are kept separate from the cross-arm validation cohort (or the external test portions). To ensure the optimal performance of the tree-based models, we selected a final parameter set based on their performance across all validation portions for each experiment. Subsequently, we re-optimised the models for each training split within the discovery cohort using this identified set of parameters. As each split would have a separate model, we ensembled (averaged) their DRSs on their corresponding intra-arm validation portions and cross-arm validation portions to obtain a more objective set of final DRSs. Furthermore, for each experiment, the threshold for all KM curves is the average of the median DRSs obtained from the validation portion of each split within the discovery cohort. This entire process is summarized in **Extended Data Fig. 1**.

For our experiments, the discovery cohort was randomly sampled into 25 different splits of train/validation/intra-arm validation (or train/validation/testing) portions (60/20/20). And to optimise the tree-based models, we conducted a random hyper-parameter search across a range of values defined in **Supplementary Table 1**, with 4096 sampling points. The parameter set that achieved the best performance on all validation splits was then chosen for subsequent inference on the corresponding intra-arm validation portions and the cross-arm validation cohort. We evaluated the performance of the models using the Concordance Index (C-Index) metric.

For the experiment in <u>**identifying responder**</u>, we trained a tree-based model using ARM3 as discovery cohort according to the strategy in described in this section. We reported the C-Index of this model on the discovery cohort in **Extended Data Table 1**. In this setting, ARM1 data were treated as cross-arm validation data. We then report the KM stratification using the DRSs from the intra-arm validation portions in **Fig. 4a**.

For the experiment in <u>**identifying patients benefitted from immunotherapy**</u>, we trained a tree-based model using ARM1 as discovery cohort while keeping ARM3 as the cross-arm validation portion according to the strategy described in this section. We reported the C-Index of this model on the discovery cohort in **Extended Data Table 1**. We then report the KM stratification using the DRSs from the intra-arm validation and the cross-arm validation portions in **Fig. 4b**. We once again stress that the KM curves for the cross-arm validation portions utilised a same cut-off to stratify the risk scores into High-Risk and Low-Risk groups.

## Data availability

The data is private and is not available to the public.

## Code availability

The code is private and is not available to the public.

## Acknowledgements

This project has been supported by Frances and Augustus Newman Foundation. QDV is funded by The Royal Marsden NHS Foundation Trust. NR and SR are part of the PathLAKE digital pathology consortium, which is partly funded from the Data to Early Diagnosis and Precision Medicine strand of the governments Industrial Strategy Challenge Fund, managed and delivered by UK. Research and Innovation (UKRI). NR and SR are also funded by the European Research Council (funding call H2020 IMI2-1133RIA). NR was also supported by the UK Medical Research Council (grant award MR/P015476/1), Royal Society Wolfson Merit Award and the Alan Turing Institute.

## Author contributions

Supervision: NR, SR and DC. Data Acquisition: CF, AG, TL, DR, KvL. Clinical input: CF, AG, TL, KvL. Methods development: QDV, Results and Analysis: QDV. Writeup: All authors.

## Competing Interests

# Supplementary

## S1. Reading and Interpreting SHAP Analysis

On a population level, a feature with lower values typically results in a lower DRS when there is either a long tail or a cluster of uniform blue dots on the left side of the plot. However, this relationship may not be consistent for all patients, as the complexity of the TiME and limitations in the dataset make it challenging to isolate the effects of individual features. For example, some patients may have high DRSs even though, on average, lower amounts of such feature tend to be associated with lower DRSs. This is evident in the figure where a feature has a long tail of blue dots on the left but also a cluster of dots with mixed red and blue colours on the right side. Due to this complexity, we subsequently only describe the most apparent population-level trend. Any other cell phenotypes or direction of effect not mentioned are assumed to be interdependent on the population level (i.e., higher is better does not mean lower is better in general).

## S2. Hyperparameter search for XGBoost

We performed a random search over the XGBoost hyperparameters to select the best model. This search space is defined as follows:

**Supplementary Table 1. Hyperparameter space when performing Random Search.** We provide the name of the parameter, as used in the Python implementation **here**, along with the range of values that we randomly sample from.

| Parameter Name | Value Ranges |
|---|---|
| num_boost_round | 8 to 256 |
| learning_rate | 0.001 to 0.1 |
| max_depth | 1 to 16 |
| subsample | One of [0.3, 0.4, 0.5, 0.6, 0.7, 0.8] |
| colsample_bytree | One of [0.3, 0.4, 0.5, 0.6, 0.7, 0.8] |
| colsample_bylevel | One of [0.3, 0.4, 0.5, 0.6, 0.7, 0.8] |
| colsample_bynode | One of [0.3, 0.4, 0.5, 0.6, 0.7, 0.8] |
| min_child_weight | 0.01 to 3.0 |
| reg_lambda | 0.1 to 2.0 |
| reg_alpha | 0.1 to 2.0 |
| booster | "booster" or "dart" |
| rate_drop | 0.1 to 0.7 |

# S3. Pathological Characteristics

**Supplementary Table 2. Pathological Characteristics.** We provide a baseline characteristics table including age, gender, primary site of disease, tumour and nodal status for patients utilised in this study.

|  | A3 N=105 | |
|---|---|---|
|  | N | % |
| **Age** | | |
| Mean (SD) | 65 | 9.5 |
| Median (IQR) | 67 | (58 to 72) |
| Range | 38 | 85 |
| **Age group** | | |
| <65 | 40 | 38 |
| 65 and above | 65 | 62 |
| **Gender** | | |
| Female | 21 | 20 |
| Male | 84 | 80 |
| **Primary Site** | | |
| OG Junction | 31 | 30 |
| Oesophagus | 45 | 43 |
| Stomach | 29 | 28 |
| **T-stage** | | |
| T0 | 1 | 1 |
| T1 | 2 | 2 |
| T2 | 8 | 8 |
| T3 | 61 | 58 |
| T4 | 28 | 27 |
| Tx | 5 | 5 |
| **N-stage** | | |
| N0 | 10 | 10 |
| N1 | 30 | 29 |
| N2 | 33 | 31 |
| N3 | 24 | 23 |
| Nx | 8 | 8 |